\documentclass[10pt,letterpaper]{article}
\usepackage[top=0.75in, left=0.75in, right=0.75in, bottom=.75in]{geometry}

\usepackage{amsmath,amssymb}

\usepackage{changepage}

\usepackage[utf8x]{inputenc}

\usepackage{textcomp,marvosym}

\usepackage{cite}

\usepackage{nameref,hyperref}

\usepackage{microtype}
\DisableLigatures[f]{encoding = *, family = * }

\usepackage[table]{xcolor}

\usepackage{array}

\newcolumntype{+}{!{\vrule width 2pt}}

\newlength\savedwidth

\usepackage[aboveskip=1pt,labelfont=bf,labelsep=period,justification=raggedright,singlelinecheck=off]{caption}

\makeatletter
\renewcommand{\@biblabel}[1]{\quad#1.}
\makeatother

\usepackage{lastpage,fancyhdr,graphicx}

\DeclareMathOperator*{\argmin}{arg\,min}

\DeclareMathOperator*{\supremum}{supremum}

\title{Deep learning as a tool for neural data analysis: speech classification and cross-frequency coupling in human sensorimotor cortex}

\author{Jesse A. Livezey\textsuperscript{1,2},
	Kristofer E. Bouchard\textsuperscript{1,2*\dag},
	Edward F. Chang\textsuperscript{3,4,5*\ddag}}

\begin{document}
\maketitle
\vspace*{0.2in}

\begin{flushleft}
\bigskip
\textbf{1} Biological Systems and Engineering Division, Lawrence Berkeley National Laboratory, Berkeley, California, USA
\\
\textbf{2} Redwood Center for Theoretical Neuroscience, University of California, Berkeley, Berkeley, California, USA
\\
\textbf{3} Department of Neurological Surgery and Department of Physiology, University of California, San Francisco, San Francisco, California, USA
\\
\textbf{4} Center for Integrative Neuroscience, University of California, San Francisco, San Francisco, California, USA.
\\
\textbf{5} UCSF Epilepsy Center, University of California, San Francisco, San Francisco, California, USA
\\
\bigskip

%
%
* These authors contributed equally to this work.




\dag kebouchard@lbl.gov

\ddag edward.chang@ucsf.edu

\end{flushleft}
\section*{Abstract}
A fundamental challenge in neuroscience is to understand what structure in the world is represented in spatially distributed patterns of neural activity from multiple single-trial measurements. This is often accomplished by learning a simple, linear transformations between neural features and features of the sensory stimuli or motor task. While successful in some early sensory processing areas, linear mappings are unlikely to be ideal tools for elucidating nonlinear, hierarchical representations of higher-order brain areas during complex tasks, such as the production of speech by humans. Here, we apply deep networks to predict produced speech syllables from cortical surface electric potentials recorded from human sensorimotor cortex. We found that deep networks had higher decoding prediction accuracy compared to baseline models, and also exhibited greater improvements in accuracy with increasing dataset size. We further demonstrate that deep network's confusions revealed hierarchical latent structure in the neural data, which recapitulated the underlying articulatory nature of speech motor control. Finally, we used deep networks to compare task-relevant information in different neural frequency bands, and found that the high-gamma band contains the vast majority of information relevant for the speech prediction task, with little-to-no additional contribution from lower-frequencies. Together, these results demonstrate the utility of deep networks as a data analysis tool for neuroscience.



\section*{Introduction}
A central goal of neuroscience is to understand what and how information about the external world (e.g., sensory stimuli or behaviors) is present in spatially distributed, dynamic patterns of brain activity. At the same time, neuroscience has been on an inexorable march away from the periphery (e.g., the retina, spinal cord), seeking to understand higher-order brain function (such as speech). The methods used by neuroscientists are typically based on simple linear transformations, which have been successful predictors in early processing stages of the nervous system for simple tasks~\cite{theunissen2000,carandini2005,schwartz2006}. However, linear methods are limited in their ability to represent complex, hierarchical, nonlinear relationships~\cite{poggio1990}, which are likely present in the neural activity of higher-order brain areas.

Multilayer deep networks can combine features in nonlinear ways when making predictions. This gives them more expressive power in terms of the types of mappings they can learn, at the cost of more model hyperparameters, more model parameters to train, and more difficult training dynamics \cite{larochelle2009}. Together with the recent success of deep learning in a number of fields including computer vision, text translation, and speech recognition~\cite{he2016, bahdanau2014, amodei2016}, the ability of deep networks to learn nonlinear function from data motivates their use for understanding neural signals. The success of deep learning in classic machine learning tasks has spurred a growth of applications into new scientific fields. Deep networks have recently been applied as classifiers for diverse types of physiological data including electromyographic (EMG), electroencephalographic (EEG), and spike rate signals~\cite{wulsin2011, stober2014, wand2014, supratak2014}, on stimulus reconstruction in sensory regions using electrocorticography (ECoG)~\cite{yang2015}, as models for sensory  and motor systems \cite{zipser1988, yamins2014, agarwal2015,mcintosh2016, benjamin2017}. While these studies have demonstrated the superior performance of deep networks as black-box predictors, the utilization of deep networks to gain understanding into brain computations is rare.

Vocal articulation is a complex task requiring the coordinated orchestration of several parts of the vocal tract (e.g., the larynx, tongue, jaw, and lips). To study the neural basis of speech requires monitoring cortical activity at high spatio-temporal resolution (on the order of tens of milliseconds) over large areas of sensorimotor cortex ($\sim$1300mm$^2$) ~\cite{bouchard2013}. Electrocorticography (ECoG) is an ideal method to achieve the simultaneous high-resolution and broad coverage requirements in humans. Using such recordings, there has been a surge of recent efforts to understand the cortical basis of speech production \cite{bouchard2013, bouchard2014, bouchard2014a, mugler2014, lotte2015, mugler2017}. For example, analyzing mean activity, Bouchard et. al.~\cite{bouchard2013} demonstrated, much in the spirit of Penfield's earlier work~\cite{penfield1937}, that the ventral sensorimotor cortex (vSMC) has a spatial map of articulator representations (i.e. lips, jaw, tongue, and larynx) that are engaged during speech production. Additionally, it was found that spatial patterns of activity across the vSMC network (extracted from trial average activity with principal components analysis at specific time points) organized phonemes along phonetic features emphasizing the articulatory requirements of production.

Understanding how well cortical surface electrical potentials (CSEPs) capture the underlying neural processing involved in speech production is important for revealing the neural basis of speech and improving speech decoding for brain-computer interfaces \cite{guenther2009,leuthardt2011}. Previous studies have used CSEPs and linear or single layer models to predict speech categories \cite{kellis2010, pei2011, mugler2014, herff2015, ramsey2017}, or continuous aspects of speech production (e.g., vowel acoustics or vocal tract configurations)~\cite{bouchard2014a, mugler2017}, with some success. However, given the challenge of collecting large number of samples across diverse speech categories, it is not clear that we should expect high performance from deep networks for speech classification. Exploring the use of deep networks to maximally extract information for speech prediction is not only important for identifying cortical computations, but also for brain machine interfaces to restore communication capabilities to humans who are "locked-in".

In general, understanding information content across neural signals, such as different frequency components of CSEPs, is an area of ongoing research \cite{crone1998a, crone1998b, buzsaki2012, michalareas2016, richter2017}. A number of studies have found relationships between different frequency components in the brains electrical potentials. These can take the form of phase and amplitude structure of beta ($\beta$) waves \cite{rubino2006, takahashi2015} or correlations between lower frequency oscillations and spiking activity or high gamma (H$\gamma$) activity~\cite{canolty2010, richter2017}. One observation is that $\beta$ band (14-30Hz) amplitude and coherence~\cite{crone1998a, pfurtscheller1999} often decreases during behavior, when the state is changing \cite{engel2010}. This has lead to the interpretation that $\beta$ may be serving a ``maintenance of state" function. However, often these effects are not differentiated between functional areas that are active versus inactive during behavior. Indeed, in other contexts, aggregation has been shown to mask structure in neural signals \cite{latimer2015}. The somatotopic organization of speech articulator control in human vSMC, and the differential engagement of these articulators by different speech sounds, potentially provides the opportunity to disentangle these issues. Furthermore, classifying behaviors, such as speech, from CSEPs can be used as a proxy for information content in a signal, obfuscating the interpretation of the results. However, this is often done using linear methods, which may not be able to take full advantage of the information in a signal. Since deep networks are able to maximize classification performance, they are an ideal candidate for comparing information content across neural signals.

In this work, we investigated deep networks as a data analytics framework for systems neuroscience, with a specific focus on the uniquely human capacity to produce spoken language. First, we show that deep networks achieve superior classification accuracy compared to linear models, with increased gains for increasing task complexity, and improved efficiency as a function of data set size. We then ‘opened the black box’ and used the deep network confusions to reveal the latent structure learned from single trials, which revealed a rich, hierarchical organization of linguistic features. Since deep networks classified speech production from H$\gamma$ activity with higher accuracy that other methods, they are also candidates for determining the relative information content across neural signals. We explored the cross-frequency amplitude-amplitude structure in the CSEPs and discovered a novel signature of motor coordination in $\beta$-H$\gamma$ coupling. Using deep networks, we then show that although there is information relevant to speech production in the lower frequency bands, it is small compared to H$\gamma$. Critically, the lower frequency bands do not add significant additional information about speech production about and beyond H$\gamma$. Furthermore, the correlations are not clearly related to overall information content and improvements in accuracy. Together, these results demonstrate the utilization of deep networks not only as an optimal black-box predictor, but as a powerful data analytics tool to reveal the latent structure of neural representations, and understanding the information content of different neural signals. 

\section*{Materials and methods}
\subsection*{Experimental data}
The experimental protocol, collection, and processing of the data examined here have been described in detail previously~\cite{bouchard2013,bouchard2014,bouchard2014a}. Briefly, four native English speaking human subjects underwent chronic implantation of a subdural electrocortigraphic (ECoG) array over the left hemisphere as part of their clinical treatment of epilepsy. The subjects gave their written informed consent before the day of surgery. The subjects read aloud consonant-vowel (CV) syllables composed of 19 consonants followed by one of three vowels (/a/, /i/ or /u/), for a total of 57 potential consonant-vowel syllables. Subjects did not produce each CV in an equal number of trials or produce all possible CVs. Across subjects, the number of repetitions per CV varied from 10 to 105, and the total number of usable trials per subject was S1 = 2572, S2 = 1563, S3 = 5207, and S4 = 1422. CVs for which there was not enough data to do cross-validation (fewer than 10 examples) were excluded per-subject.

\subsection*{Signal processing}
Cortical surface electrical potentials (CSEPs) were recorded directly from the cortical surface with a high-density (4mm pitch), 256-channel ECoG array and a multi-channel amplifier optically connected to a digital signal processor (Tucker-Davis Technologies [TDT], Alachua, FL). The time series from each channel was visually and quantitatively inspected for artifacts or excessive noise (typically 60 Hz line noise). These channels were excluded from all subsequent analysis and the raw CSEP signal from the remaining channels were then common-average referenced and used for spectro-temporal analysis. For each useable channel, the time-varying analytic amplitude was extracted from 40 bandpass filters (Gaussian filters, logarithmically increasing center frequencies and semi-logarithmically increasing band-widths) with the Hilbert transform. This amplitude for each filter band was z-scored to a baseline window defined as a period of time in which the subject was silent, the room was silent, and the subject was resting. For each of the bands defined as: theta [4-8 Hz], alpha [9-14 Hz], low beta [15-20 Hz], high beta [21-29 Hz], gamma [30-59 Hz], and high gamma [75-150 Hz], individual bands from the 40 Gaussian bandpassed amplitudes were grouped and averaged according to center frequencies. For the use in deep networks, this signal was down-sampled to a rate such that the center frequency-to-sampling rate ratio was constant (ratio$=112.5/200$) per frequency component. Based on previous results~\cite{bouchard2013, bouchard2014, bouchard2014a}, we focused on the electrodes in the ventral sensorimotor cortex (vSMC). The activity for each of the examples in our data set was aligned to the acoustic onset of the consonant-to-vowel transition. For each example, a window $0.5$ seconds preceding and $0.8$ seconds following the acoustic onset of the consonant-to-vowel transition was extracted. The mean of the first and last $\sim 4\%$ time samples was subtracted from the data per electrode and trial (another form of amplitude normalization that is very local in time).
This defined the z-scored amplitude that is used for subsequent analysis.

\subsection*{Deep networks}\label{deepnet_training}
Supervised classification models often find their model parameters, $\hat\Theta$, which minimize the negative log-likelihood of the training data and labels, $\{x^{(i)}, y^{(i)}\}$, under a model which gives the conditional probability of the labels given the input data
\begin{equation}
\hat \Theta = \argmin_\Theta\ -\log P(Y|X; \Theta),\ \{x^{(i)}, y^{(i)}\}.
\end{equation}
Deep networks typically parametrize this conditional probability with a sequence of linear-nonlinear operations. Each layer in a fully-connected network consists of an affine transform followed by a nonlinearity: 
\begin{equation}\label{fc}
\begin{split}
h^1 &= f(w^1\cdot x+b^1),\\
h^i &= f(w^i\cdot h^{i-1}+b^i),\ \text{with} \\
\Theta &= \{w_1,\ldots, w_n, b_1,\ldots, b_n\}
\end{split}
\end{equation}
where $x$ is a batch of input vectors, $w^i$ and $b^i$ are trainable parameters (weights and biases, respectively) for the $i$th layer, $h^i$ is the $i$th hidden representation, and $f(\cdot)$ is a nonlinearity which can be chosen during hyperparameter selection. Single layer classification methods, such as multinomial logistic regression, are a special case of deep networks with no hidden representations and their corresponding hyperparameters.

For the fully-connected deep networks used here, the CSEP features were rasterized into a large feature vector per-trial in a window around CV production. These feature vectors are the input into the first layer of the fully connected network. The final layer non-linearity is chosen to be the softmax function:
\begin{equation}
	P(\hat y_i) = \text{softmax}(h_i) = \frac{\exp(h_i)}{\sum_j \exp(h_j)}
\end{equation}
where $h_i$ is the $i$th element of the hidden representation. This nonlinearity transforms a vector of real numbers into a vector which represents a one-draw multinomial distribution. It is the negative log-likelihood of this distribution over the training data which is minimized during training.

To train and evaluate the networks, the data is organized into 10 groupings (folds) with mutually exclusive validation and test sets and 80-10-10\% splits (training-validation-testing). Since some classes may have as few as 10 examples, it was important to split each class proportionally so that all classes were equally distributed. Training terminated when the validation accuracy did not improve for 10 epochs and typically lasted about 25 epochs. Theano, Pylearn2, and Scikit-learn~\cite{pedregosa2011, goodfellow2013, theano2016} were used to train all deep and linear models.

As baseline models, we trained multinomial logistic regression using the same hyperparameter selection method as deep networks (described below) with no hidden layers. These models had the highest classification accuracy compared to other linear classifiers, i.e. linear support vector machines and linear discriminant analysis.

\subsubsection*{Hyperparameter search}
Deep networks have a number of hyperparameters that govern network architecture and optimization, such as the number of layers, the layer nonlinearity, and the optimization parameters. The full list of hyperparameters is listed in \nameref{S1_Appendix}.

For all results, hyperparameters were selected by choosing the model with the best mean validation classification accuracy across 10 folds. Hyperparameter search was done using random search \cite{bergstra2012}. Since our datasets were relatively small for training deep networks, we regularized the models in three ways: dropout, weight decay, and filter norm-clipping in all layers of the model. The dropout rate, activation-rescaling factor, max filter norm, and weight decay coefficient were all optimized hyperparameters.

\subsubsection*{Classification tasks}
Each subject produced a subset of the 57 CV and the classification methods were trained to predict the subset. Each CV can also be classified as containing 1 of 19 consonants or 1 of 3 vowels. Similarly, a subset of the constants can be grouped into 1 of 3 vocal tract constriction location categories or 1 of 3 vocal tract constriction degree categories. The model CV predictions were then tabulated within these restricted labelings. 

As there are drastically different numbers of classes between the different tasks, as well as subtle differences between subjects, classification accuracies and changes in accuracies are all normalized to chance. In each case, chance accuracy is estimated by assuming that test set predictions are drawn randomly from the training set distribution. This process was averaged across 100 random resamplings per fold, training fraction, subject, etc. Estimating chance accuracy by training models on data with shuffled labels was not possible for consonant constriction location and degree tasks since not all CVs were used and occasionally networks would predict 0 trials within the task which would give undefined chance accuracy.

\subsubsection*{Information content in neural signals}

For a given experimentally defined behavior, such as CV speech production, the information about the task is presumably present in the activity of the brain, which we coarsely measure with different frequency components of the recorded CSEPs. The information about the task in the measurements can be formalized by the mutual information between the task variable $Y$ and the neural measurement variable $X$~\cite{cover2012}
\begin{equation}
I(Y;X) = \sum_{y, x}P(y, x)\log_2\left( \frac{P(y, x)}{P(y)P(x)}\right).
\end{equation}
It is not possible to calculate this quantity directly because we do not know the joint distribution of neural measurements and speech tokens, $P(X, Y)$ and cannot easily approximate it due to the small number of samples ($\sim10^3$) compared to the dimensionality of each measurement ($\sim10^4$). However, we can classify the behavior from the neural data using statistical-machine learning methods, i.e. deep learning. For a supervised classification task, machine learning methods typically generate conditional probabilities $P(\hat Y | X)$. Since we know the ground-truth behavior for each measurement, we can use the classifier to compute the mutual information between the behavior state, $Y$, and the predicted state, $\hat Y$
\begin{equation}
I(\hat Y;Y) = \sum_{y, \hat y}P(\hat y| y)P(y)\log_2\left( \frac{P(\hat y| y)P(y)}{P(y)P(\hat y)}\right).
\end{equation}
The data processing inequality tell us that this quantity is a lower bound to $I(Y;X)$.

Given this lower bound, if everything else is held constant, the classification method with highest accuracy will lead the tightest estimate of the mutual information between the task and neural data, $I(Y;X)$, which is a quantity that is relevant for future experimental hardware, methods, and data preprocessing development.

This quantity is closely related to a second measure of classifier performance, the Channel Capacity (CC). To compare our results with previous speech classification studies, we report estimated CC, which is measured in bits per symbol, in addition to classification statistics. CC is a unified way of calculating the effectiveness of different speech classifiers, which can have differing numbers of classes and modalities. The channel capacity, $CC$, between the ground truth class, $Y$, and predicted class, $\hat Y$, is defined as:
\begin{equation}\label{eq:capacity}
CC = \supremum_{P(Y)}I(\hat Y; Y)= \supremum_{P(Y)}\sum_{\hat y_i, y_j}P(\hat y_i| y_j)P(y_j)\log_2\left(\frac{P(\hat y_i| y_j)}{P(\hat y_i)}\right).
\end{equation}

For previous work, we must approximate the channel capacity since we do not have access to the details of the classification performance, $P(\hat Y | Y)$. Wolpaw et. al.~\cite{wolpaw2002} suggest an approximation that assumes all classes have the same accuracy as the mean accuracy and all errors are distributed equally (note that this second assumption is generally not true in speech, i.e. Fig~\ref{fig:hierarchy}C, also noted in \cite{mugler2014}). To make a fair comparison, we compute this approximate value for our results in addition to the exact value. For our data, we find that the approximation underestimates the true channel capacity for the CV and consonant task. The Information Transfer Rate (ITR) is also commonly reported, which is the channel capacity divided by the symbol duration in time. Since we are considering fixed length measurements (1.3 s), we report channel capacity rather than ITR.

\subsubsection*{Classification performance scaling with training data size}
We compared performance scaling of different models by training on different fractions of the training set. For each fraction of the data, each class was subsampled individually to ensure all classes were present in the training set. The aggregate slopes were calculated with ordinary least-squares regression. The validation and test sets were not subsampled. Hyperparameters were chosen independently for each fraction of the training data.

\subsection*{Structure of deep network predictions}
Neuroscientists commonly study the model/confusions of linear analysis methods to gain insight into the structure of neural data. Deep networks can learn high dimensional, nonlinear features from data. Here, these features are learned by training the networks to perform classification, i.e. maximize $P(\hat Y_i|X_i; \Theta)$ where the subscript $i$ indicates true class membership. It has been shown that these features contain more information than the thresholded multinomial classification prediction \cite{warde2014,hinton2015}. The off-diagonal values: $P(\hat Y_i|X_j; \Theta)$, $i\neq j$, in this learned distribution represent prediction uncertainty for a given CSEP measurement.  Uncertainty is learned during the training process and larger pairwise uncertainty between class labels means that the model has a harder time distinguishing those classes, i.e. the neural data for those class labels in more similar.

To gain insight into the nature the vSMC neural activity, we analyzed the structure of deep network predictions. The mean network prediction probabilities on the test set are used as features for each CV. A dendrogram was computed from the hierarchical clustering (Ward's method) of these features. A threshold in the cluster distance was chosen by visual inspection of the number of clusters as a function of distance, and the linguistic features were labeled by hand. The CV order from this clustering was used to order the features in the soft-confusion matrix and accuracy per CV. The soft confusion matrix shows mean network prediction probabilities on the test set rather than the aggregated thresholded predictions often shown in confusion matrices. 

To compare the articulatory features and the deep network features quantitatively across subjects, pairwise distances between CVs were computed in both the articulatory and deep network spaces (see \nameref{S1_Fig} for articulatory features). These pairwise distances were then correlated per for each CV and subject and articulatory grouping.

\subsection*{Cross-band amplitude-amplitude correlations} 
To examine the relationship between the amplitudes of different frequency components of recorded CSEPs, we first performed a correlation analysis. For this analysis, the data was trial averaged per CV then organized into a data-tensor, $D_{\text{CV},\text{frequency},\text{electrode},\text{time}}$. The frequency bands were then either used individually or aggregated into canonical frequency components, such as H$\gamma$
\begin{equation}
\bar D(\text{H}\gamma)_{\text{CV},\text{electrode},\text{time}} =  \langle \bar D_{\text{CV},\text{frequency},\text{electrode},\text{time}} \rangle_{\text{frequency} \in \text{H}\gamma}.
\end{equation}

$\bar D(\text{H}\gamma)_{\text{CV},\text{electrode},\text{time}}$ was correlated across time at 0 ms lag with each of the 40 Gaussian bandpassed amplitudes averaged across CVs and electrodes. The correlation between $\bar D(\text{H}\gamma)_{\text{CV},\text{electrode},\text{time}}$ and $\bar D(\beta)_{\text{CV},\text{electrode},\text{time}}$ was computed and histogrammed across CVs and electrodes. The average H$\gamma$ power was averaged in a window 70 ms before and 140 ms after the CV acoustic transition and histogrammed across CVs and electrodes. This window was chosen as it is the most active and informative time period for consonants and vowels.

\subsection*{Resolved cross-band amplitude-amplitude correlation}
For a band, $\mathbb{B}$ and H$\gamma$, the $\mathbb{B}$-H$\gamma$ amplitude-amplitude correlation and average H$\gamma$ power with positive average amplitude (greater than baseline) were fit with linear regression. The electrodes were then divided into ``active" and ``inactive" per CV by thresholding the average H$\gamma$ amplitude where the linear fit predicted 0 correlation.

The active and inactive electrodes per CV were separated and $\bar D(\text{H}\gamma)_{\text{CV},\text{electrode},\text{time}}$ was correlated across time at 0 ms lag with each of the 40 Gaussian bandpassed amplitudes averaged across CVs and electrodes independently for both sets and for each subject.

\subsection*{Classification from other frequency bands}
An extended sets of lower frequency features per trial were used in addition to the H$\gamma$ features for each of the theta, alpha, low beta, high beta, and gamma bands. For each frequency band, fully-connected deep networks were trained first on the individual bands's features and then with the band's features concatenated with the H$\gamma$ features. Deep network training was done in the same manner at the networks trained soley on H$\gamma$ features. The resulting classification accuracies were then compared with the baseline H$\gamma$ classification accuracy and then with the band's features concatenated with the H$\gamma$ features.



\section*{Results}
\begin{figure}[!htbp]
	\begin{center}
		\includegraphics[width=6in]{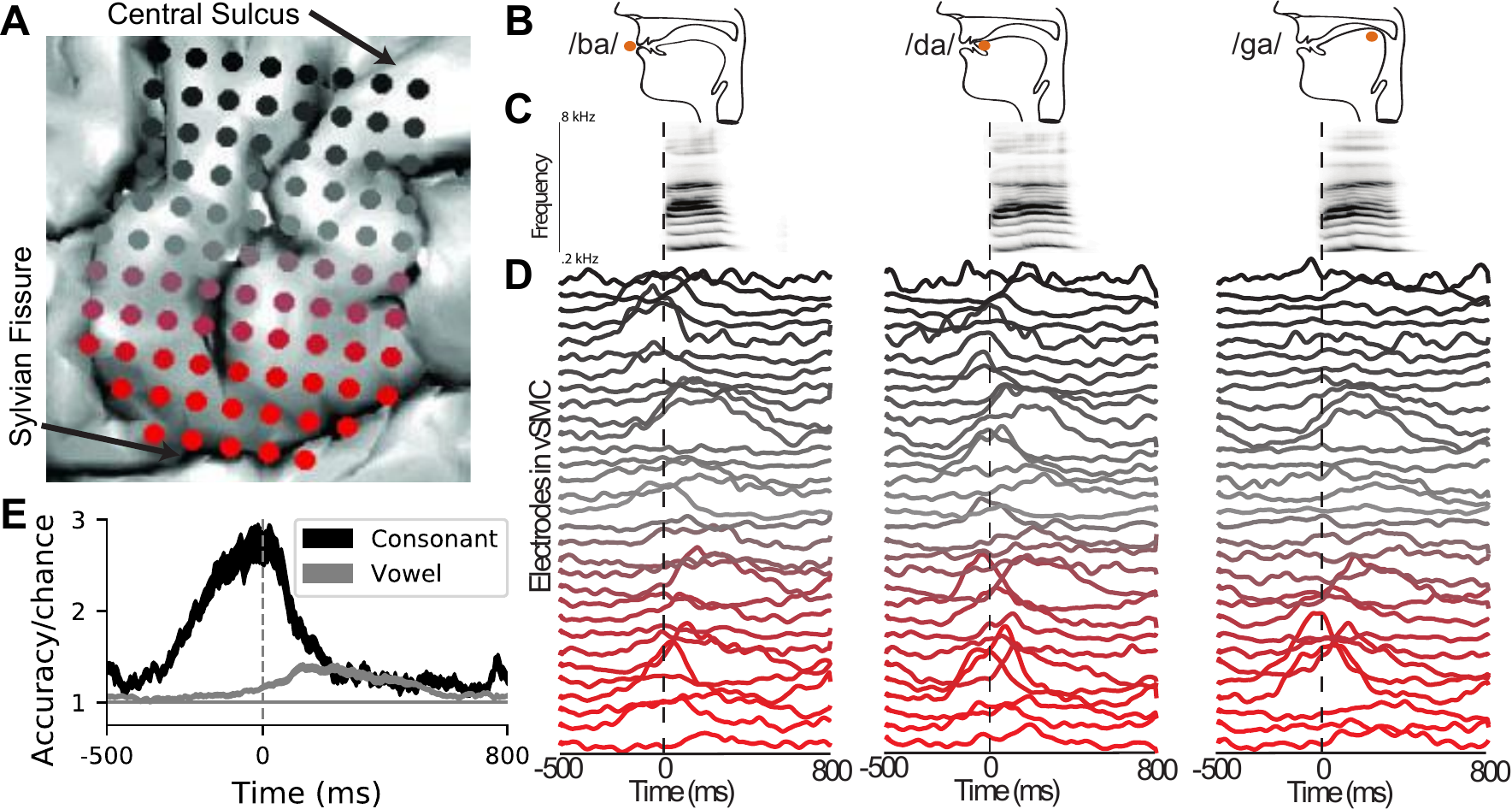}
	\end{center}
	\caption{{\bf Human ECoG recordings from ventral sensorimotor cortex (vSMC) during speech production.} \textbf{A} Electrodes overlaid on vSMC. Electrodes are colored red-to-black with increasing distance from the Sylvian Fissure. \textbf{B-D} Task and data summary for three different consonant-vowel (CV) utterances. \textbf{B} Vocal tract configuration and point of constriction (orange dot) during the consonant for the production of /ba/ (lips), /da/ (coronal tongue), and /ga/ (dorsal tongue). \textbf{C)} The audio spectrogram aligned to the consonant-to-vowel acoustic transition (dashed line). \textbf{D} Mean across trials of the H$\gamma$ amplitude from a subset of electrodes in vSMC aligned to CV transition. Traces are colored red-to-black with increasing distance from the Sylvian Fissure as in \textbf{A}. The syllables /ba/, /da/, and /ga/ are generated by overlapping yet distinct spatio-temporal patterns of activity across vSMC. \textbf{E} Logistic regression accuracy for consonants and vowels plotted against time aligned to the CV transition averaged across subjects and folds. Black and grey traces are average ($\pm$ s.e.m., $n=40$) accuracies for consonants ($18-19$ classes) and vowels (3 classes) respectively.}
	\label{fig:ecog_data}
\end{figure}

A subset of the electrodes of the ECoG grid overlaid on the vSMC of Subject 1 is shown in Fig~\ref{fig:ecog_data}A. Cortical electric surface potentials were recorded from the left hemisphere of 4 subjects during the production of a set of consonant-vowel syllables which engage different section of the vocal tract, as shown in Fig~\ref{fig:ecog_data}B, to produce acoustics which are shown in Fig~\ref{fig:ecog_data}C.  The trial-averaged z-scored high gamma (H$\gamma$) amplitude recorded during the production of the syllables from Fig~\ref{fig:ecog_data}B show spatially and temporally distributed patterns of activity (Fig~\ref{fig:ecog_data}D). Here we see that cortical surface electrical potentials recorded from vSMC during the production of CVs consists of multiple spatially and temporally overlapping patterns.

Spatiotemporal patterns of activity represent information about the produced syllables \cite{bouchard2013}. This is shown by training multinomial logistic regression models independently at each time point using all electrodes in vSMC (Fig. \ref{fig:ecog_data}E). Across subjects, the consonant classification accuracy rises from chance approximately 250 ms before the consonant-vowel acoustic transition at 0 ms, which precedes the acoustic production of the consonants, indicating the motor nature of the recordings. Consonant classification accuracy remains above chance for approximately 200 ms into vowel acoustics production. Vowel classification accuracy rises just before the transition to vowel acoustics production and remains above chance for approximately 500 ms. These results show that the consonant and vowel identity is encoded in the H$\gamma$ amplitude in partially-overlapping temporal segments.

\subsection*{Deep Learning for speech classification}

\subsubsection*{Deep networks outperform standard methods for consonant-vowel classification from high gamma amplitude}

It has been shown that CSEPs contain information about motor control \cite{bouchard2013, bouchard2014,bouchard2014a,mugler2014, herff2015}. Regressing CSEP time-frequency features onto behavioral features with linear methods has been used to elucidate the information content. Linear decoders can put a lower bound on the behaviorally relevant information in a measurement, but the restriction to linear mappings may limit the amount of information they are able to extract from the neural signal.

\begin{figure}[!htbp]
	\begin{center}
		\includegraphics[width=7in]{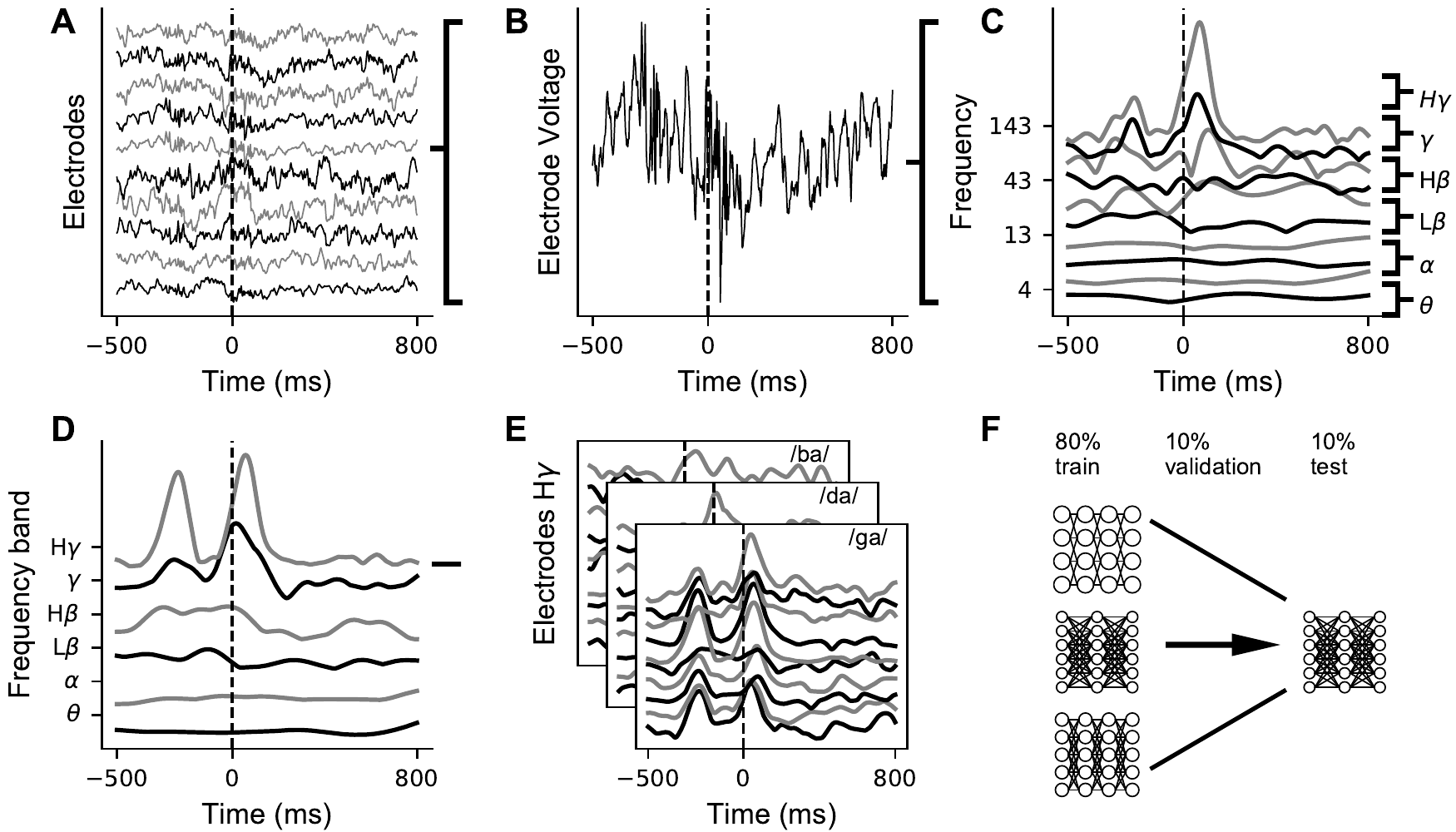}
	\end{center}
	\caption{{\bf Data processing and deep network training pipeline for ECoG data.} \textbf{A} Cortical surface electrical potentials plotted against time for a subset of the vSMC electrodes segmented to the CV production window. Electrodes have an arbitrary vertical offset for visualization. \textbf{B} Voltage for one electrode. \textbf{C} The z-scored analytic amplitude is shown for a subset of the 40 frequency ranges used in the Hilbert Transform as a function of time. \textbf{D} The 40 ranges used in the Hilbert Transform are grouped and averaged according to whether their center frequency is part of each traditional neuroscience band. \textbf{E} For a particular analysis, a subset of the bands are chosen as features, and this process was repeated for each trial (sub-pane) and electrode (trace within each sub-pane) in vSMC. Each data sample consists of one trial's H$\gamma$ activity for all electrodes in vSMC. \textbf{F} Data were partitioned 10 times into training, validation, and testing subsets (80\%, 10\%, and 10\% respectively) with independent testing subsets. We trained models that varied in a large hyper-parameter space, including network architecture and optimization parameters, symbolized by the 3 networks on the left with differing numbers of units and layers. The optimal model (right) is chosen based on the validation accuracy and results are reported on the test set.}
	\label{fig:pipeline}
\end{figure}

Deep networks can learn more complex, nonlinear mappings, which can potentially extract more information from a neural signal. Thus, they may be able to put a tighter lower bound on the information relevant for speech classification contained in CSEP features. To test this, fully connected deep networks and baseline multinomial logistic regression models were trained on z-scored $H\gamma$ amplitude from all electrodes in vSMC and time points in a window around CV production. Fig~\ref{fig:pipeline} shows how the raw CSEP measurements are preprocessed into time-frequency features across behavioral trials, selected and grouped into datasets, and are used in the deep network hyperparameter cross-validation loop. The networks with the highest validation accuracy, averaged across 10 folds, were selected and their results on a held-out test set are reported.

\begin{figure}[!htbp]
 	\begin{center}
 		\includegraphics[width=5in]{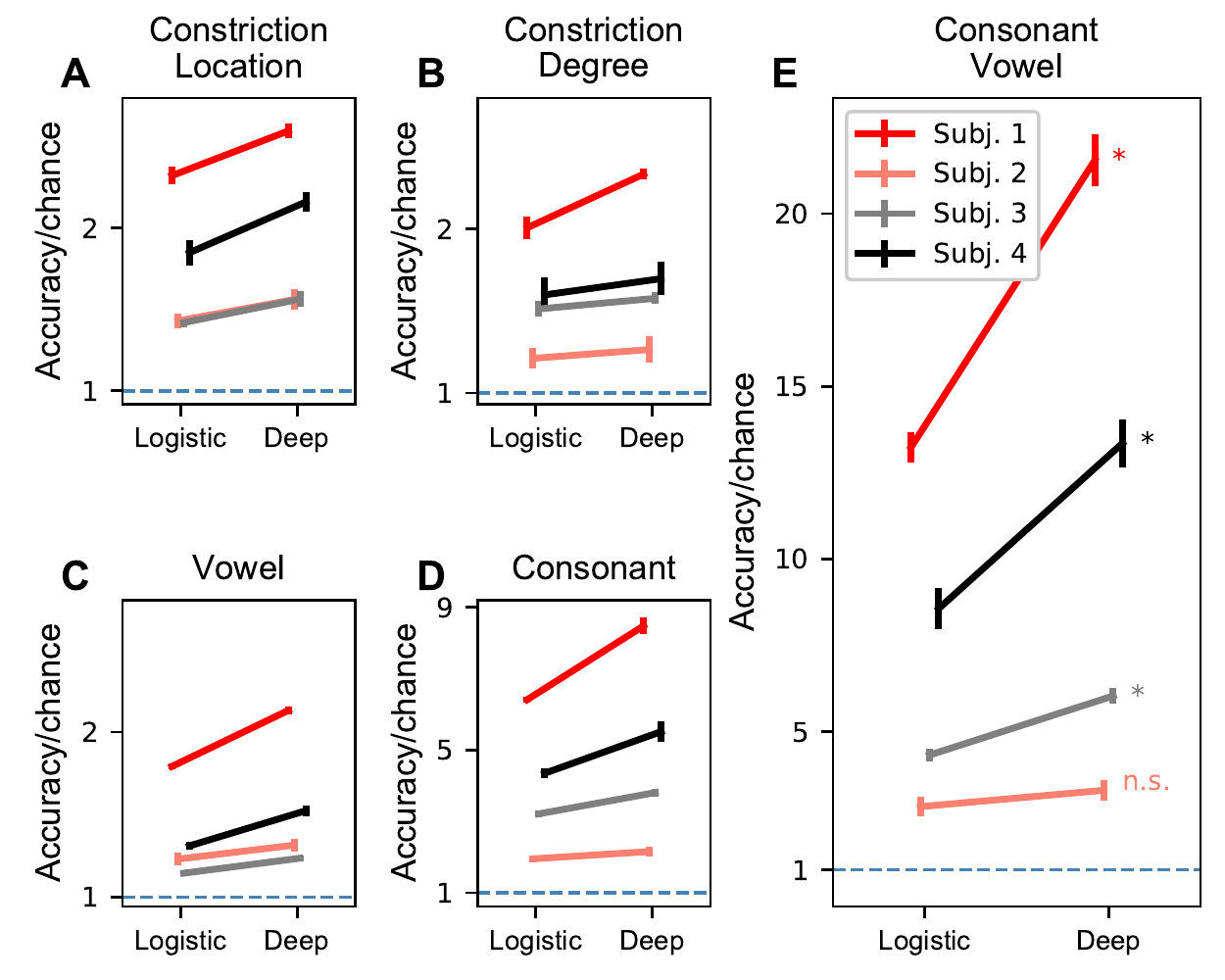}
 	\end{center}
 	\caption{{\bf Classification accuracy of logistic regression versus deep networks for different classification tasks.} For \textbf{A}-\textbf{E}, accuracies ($\pm$ s.e.m., n=10) are normalized to chance (chance = 1, dashed blue line) independently for each subject and task. Points on the left are multinomial logistic regression accuracy and are connected to the points on the right which are deep network accuracies for each subject. Subject accuracies have been left-right jittered to prevent visual overlap and demarcated with color (legend in \textbf{E}). \textbf{A-D} Classification accuracy when CV predictions are restricted to consonant constriction location (\textbf{A}), consonant constriction degree (\textbf{B}), vowel (\textbf{C}), or consonant (\textbf{D}) classification tasks. \textbf{E} Classification of entire consonant-vowel syllables from H$\gamma$ amplitude features. $^*p<0.05$, WSRT, Bonferroni corrected with $n=4$. n.s., not significant. Significance was tested between deep network and logistic regression accuracies.}
 	\label{fig:accuracy}
 \end{figure}

Behaviorally, speech is organized across multiple levels. Even within the simple CV task examined here, there are multiple levels of attributes that can be associated with each CV syllable. The simplest description of the CVs correspond to the consonant constriction location, consonant constriction degree, or vowel labels (3-way tasks). Fig~\ref{fig:accuracy}A-C shows the accuracies in these cases respectively. For these tasks, subjects with baseline accuracy close to chance see little-to-no improvement and subjects with larger improvements are limited by the complexity of the 3-way classification task. An intermediate level of complexity is the consonant label (18 or 19-way, Fig~\ref{fig:accuracy}D). The highest deep network accuracy for a single subject on the consonant task is for Subject 1 which is 44.9 $\pm$ 3.0\% (8.5 times chance, 5.3\%) and 34.0 $\pm$ 1.5\% (6.4 times chance, 5.3\%) for logistic regression which is a 32.4\% improvement. Mean consonant classification accuracy across subjects (19 way) with deep networks is 26.8 $\pm$ 12.6\%. For logistic regression, it is 21.4 $\pm$ 8.7\%.
 
Finally, the most complex task is CV classification which has between 54 and 57 classes across subjects. The highest deep network accuracy for a single subject on the CV task is for Subject 1 which is 38.3 $\pm$ 2.9\% (21.5 times chance, 1.8\%) and 23.6 $\pm$ 2.1\% (13.2 times chance, 1.8\%) for logistic regression which is a 63.1\% improvement (Fig~\ref{fig:accuracy}E). Mean CV classification accuracy across subjects (54-57 way) with deep networks is 19.9 $\pm$ 12.6\%. For logistic regression, it is 13.1 $\pm$ 7.4\%. Per subject improvements for Subjects 1 through 4 are 8.3x ($p<0.05$), 0.5x (n.s.), 1.7x ($p<0.05$), and 4.8x ($p<0.05$). For each subject, a Wilcoxon Signed-Rank Test (WSRT) was performed and the resulting p value was Bonferroni corrected ($n=4$). For the 3 significant results, the p-value was at the floor for a WSRT with $n=10$ samples and no equal differences.

The results described above contain many potential sources of variation. To test the significance of these variations, we use an ANOVA with subject, model type (deep network versus logistic regression), task complexity (CV versus consonant versus {vowel, location, degree}), and model-task complexity interaction as categorical groupings. This model is significant (f-statistic: 115.8, $p<1\times 10^{-10}$) and all coefficients were significant at $p<0.001$ with Subject 1, CV task, and logistic regression as the baseline. This shows that deep networks are able to provide better estimate of information contained in the H$\gamma$ amplitude as compared to linear methods.

The number of speech tokens, duration of a task, and recording modality often differ from study to study~\cite{wolpaw2002}. This means that quantifying the quality of speech classification from neural signals using accuracy or accuracy normalized to chance can be misleading. The Information Transfer Rate (ITR, bits per second) is a quantity that combines both accuracy and speech in a single quantity~\cite{wolpaw2002}. Since we are comparing fixed length syllables, this is equivalent to calculating the number of bits per syllable which can be calculated with the Channel Capacity (CC, Eq. \ref{eq:capacity}). The ITR can be calculated by diving the CC by the syllable duration. A summary of the accuracy results along with channel capacity estimates are summarized in Table \ref{table:results} and compared against the results of Mugler et al.~\cite{mugler2014} which has a similar task and used linear discriminant analysis (LDA) as the classifier. Deep networks achieve state of the art classification accuracy and have the highest CC, and therefore ITR, on the full CV task. The state of the art accuracy and ITR are important quantities for brain-computer interfaces, which often limit communication rates in clinical applications.

\begin{table}[!htbp]
	\caption{Classification and Channel Capacity Results}
	\label{table:results}
	\begin{center}
		\begin{tabular}{llll}
			\multicolumn{1}{c}{\textbf{Model}}  &\multicolumn{1}{c}{\textbf{Accuracy}} &\multicolumn{1}{c}{\textbf{Acc./Chance}} &\multicolumn{1}{c}{\textbf{CC (Bits/Syllable)}}
			\\ \hline \\
\textbf{Deep network, 57 CV, single subj.}         &\textbf{38.3 $\pm$ 2.9\%} & \textbf{21.5x} & \textbf{1.3 (3.09 exact)} \\
Deep network, 57 CV, subj. average        &19.9 $\pm$ 12.6\% & 11.1x & 0.53 (2.18 exact) \\
Logistic Regression, 57 CV, single subj.  &23.6 $\pm$ 2.1\% & 13.2x & 0.61 (2.47 exact) \\
Logistic Regression, 57 CV, subj. average &13.1 $\pm$ 7.4\% & 7.2x & 0.26 (1.91 exact) \\
\hline
\textbf{Deep network, 19 cons., single subj.}         &\textbf{44.9 $\pm$ 3.0\%} & \textbf{8.5x} & \textbf{0.99 (1.9 exact)} \\
Deep network, 19 cons., subj. average        &26.8 $\pm$ 12.6\% & 5.0x & 0.43 (1.03 exact) \\
Logistic Regression, 19 cons., single subj.  &34.0 $\pm$ 1.5\% & 6.4x & 0.59 (1.35 exact) \\
Logistic Regression, 19 cons., subj. average &21.4 $\pm$ 8.7\% & 4.0x & 0.27 (0.75 exact) \\
LDA~\cite{mugler2014}, 24 cons., single subj.  &36.1\% & 4.9x & 0.75 \\
LDA~\cite{mugler2014}, 24 cons., subj. average &20.4 $\pm$ 9.8\% & 2.8x & 0.25 \\
\hline
\textbf{Deep network, 3 vowels, single subj.}         &\textbf{71.1 $\pm$ 1.9\%} & \textbf{2.1x} & \textbf{0.67 (0.44 exact)} \\
Deep network, 3 vowels, subj. average        &51.7 $\pm$ 12.1\% & 1.6x & 0.28 (0.15 exact) \\
Logistic Regression, 3 vowels, single subj.  &59.8 $\pm$ 2.1\% & 1.8x & 0.39 (0.21 exact) \\
Logistic Regression, 3 vowels, subj. average &45.7 $\pm$ 8.8\% & 1.4x & 0.16 (0.07 exact) \\
LDA~\cite{mugler2014}, 15 vowels, single subj.  &23.9\% & 1.9x & 0.22 \\
LDA~\cite{mugler2014}, 15 vowels, subj. average &19.2 $\pm$ 3.7\% & 1.5x & 0.12 \\
		\end{tabular}
	\end{center}
\end{table}

\subsubsection*{Deep networks scale better with dataset size compared to standard methods}
How the accuracy and precision of data analysis results scale with dataset size is an important metric for designing future experiments. This is especially true when working with human subjects and invasive or time consuming data collection methods. In the context of brain-computer interface (BCI) research, maximizing BCI performance is a central goal and so understanding how performance is limited by dataset size or decoding/classification methods is crucial for improving clinical use and understanding the potential role of deep networks in BCIs.

Deep networks are well known for their performance on enormous machine learning datasets. Since neural datasets are typically much smaller, we sought to explore the data efficiency of deep networks relative to linear networks in the context of speech classification from CSEPs. We subsampled the training datasets by up to 50 percent in order to estimate accuracy improvements as a function of dataset size. The subsampled training dataset sizes and resulting classification accuracies were then used to estimate the slope of the accuracy as a function of dataset size.

\begin{figure}[!htbp]
	\begin{center}
		\includegraphics[width=5in]{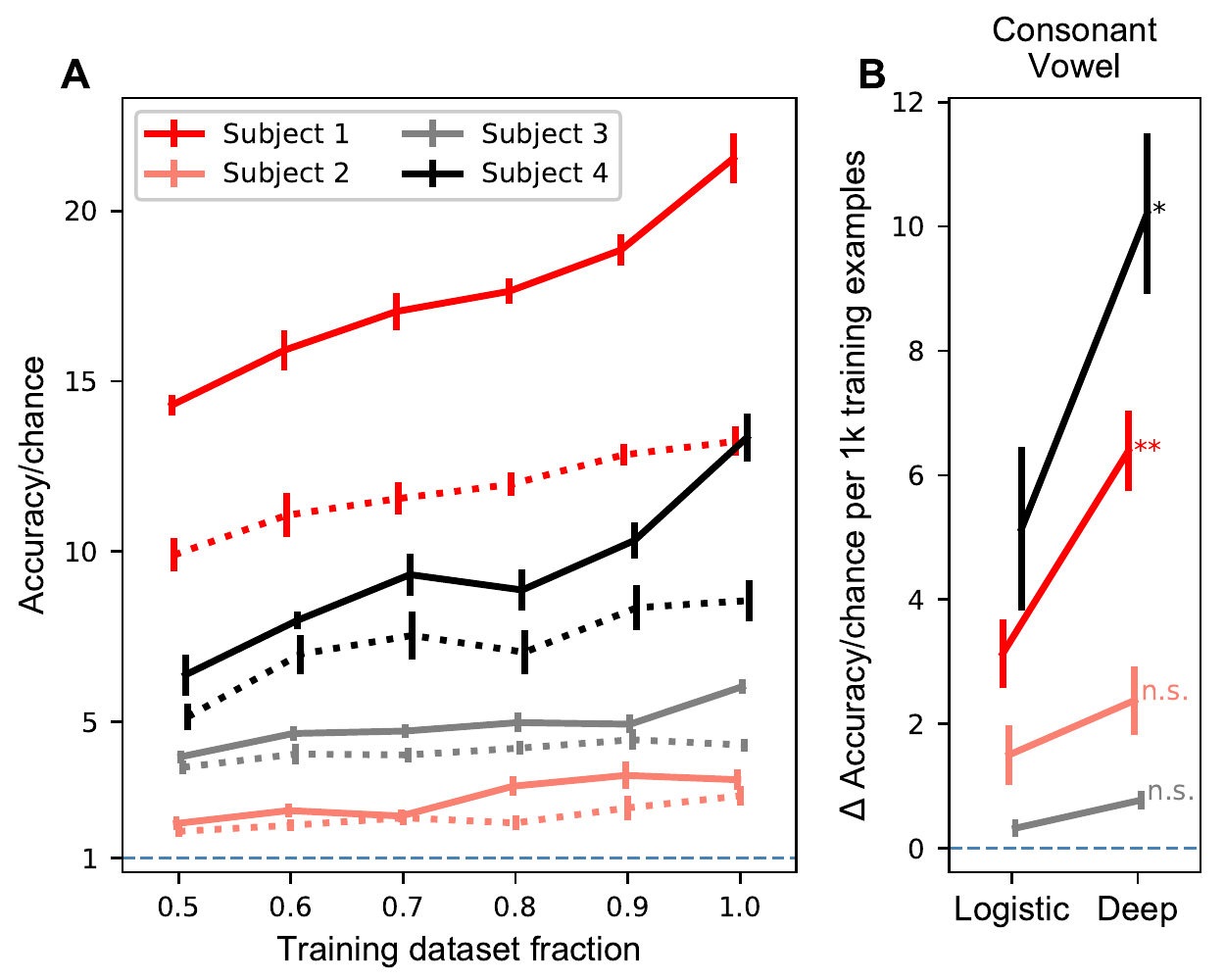}
	\end{center}
	\caption{{\bf Classification accuracy improvement as a function of training dataset size for logistic regression versus deep networks.} Accuracies ($\pm$ s.e.m., $n=10$) are normalized to chance (chance  = 1, dashed blue line) independently for each subject. Subject error bars have been left-right jittered to prevent visual overlap and demarcated with color (legend in \textbf{A}). \textbf{A} Average classification accuracy ($\pm$ s.e.m., $n=10$) normalized to chance for the CV task as a function of the fraction of training examples used for logistic regression (dotted lines) and deep networks (solid lines). \textbf{B} Change in classification accuracy normalized to chance per 1,000 training examples. The total training set sizes vary significantly between subjects so there is an additional per-subject normalization factor between the slopes in \textbf{A} and \textbf{B}. $^*p<0.05$, $^{**}p<0.001$, t-test on regression slopes between deep network and logistic regression, Bonferroni corrected with $n=4$. n.s., not significant.}
	\label{fig:slope}
\end{figure}

As the fraction of the training set was changed from 0.5 to 1.0, deep network accuracies improve (Fig.~\ref{fig:slope}A, solid lines). The accuracy and change in accuracy for deep networks is higher than for logistic regression (Fig.~\ref{fig:slope}A, dotted lines). The improvement can be estimated by fitting regression lines for each model and subject. Subjects 1 and 4 show a significant improvement per thousand training examples in the change in accuracy normalized to chance for deep networks compared to logistic regression (Fig~\ref{fig:slope}B). For the subject with highest accuracy (Subject 1), the change in accuracy over chance per 1,000 training examples for deep networks and logistic regression are 6.4x $\pm$ 0.6 and 3.1x $\pm$ 0.6 respectively. For the subject with highest slope (Subject 4), the change in accuracy over chance per 1,000 training examples for deep networks and logistic regression are 10.2x $\pm$ 1.3 and 5.1x $\pm$ 1.3 respectively. Across subjects, deep networks scale better with dataset size than logistic regression with an improvement of 4.9x $\pm$ 3.7  and 2.5x $\pm$ 1.8 over chance per 1,000 training samples respectively.

The results reported above vary across subject and model type. To test the significance of these variations, we performed an ANOVA with subject and model type (deep network versus logistic regression) as categorical groupings with no interaction. This model is significant (f-statistic: 30.7, $p<1\times 10^{-10}$) and all coefficients are significant at $p<0.001$ with Subject 1 and logistic regression as the baseline. Overall, we find that deep networks scale better than logistic regression as a function of dataset size. Additionally, for the subjects with higher accuracy, there is no indication that accuracy is saturating, which implies that accuracy of the deep networks would have continued to improve with more data collection, an important result for future studies.

\subsection*{Deep networks have classification confusions that recapitulate the articulatory organization of vSMC}

Despite being able to mathematically specify the computations happening everywhere in the model, deep networks are often described as ``black boxes". What deep networks learn and how it depends on the structure of the dataset is not generally understood. This means that deep networks currently have limited value for scientific data analysis because their learned latent structure cannot be mapped back onto the structure of the data. Many current uses of deep networks in scientific applications rely on their high accuracy and do not inspect the network computations \cite{alipanahi2015, baldi2015, benjamin2017}, although there are results in low dimensional networks~\cite{zipser1988} and early sensory areas~\cite{mcintosh2016}. Nevertheless, deep networks' ability to consume huge datasets without saturating performance means that expanding their use in science is limited by our understanding of their ability to learn about the structure of data. For the dataset consider in this work, previous studies have shown that an articulatory hierarchy can be derived from the trial-averaged H$\gamma$ amplitude using principal components analysis at hand-selected points in time \cite{bouchard2013}. Note that the articulatory structure of the consonants are not contained in the CV labels.

To explore whether deep networks can infer this latent structure from the training data, we examined the structure of network output to better understand the organization of deep network syllable representations extracted from vSMC. Deep networks used for classification predict an entire distribution over class labels for each data sample. This learned distribution has been shown to be a useful training target in addition to the thresholded class labels \cite{warde2014, hinton2015}. We clustered these learned representations and compared them to articulatory representations of the CVs.

\begin{figure}[!htbp]
	\begin{center}
		\includegraphics[width=6in]{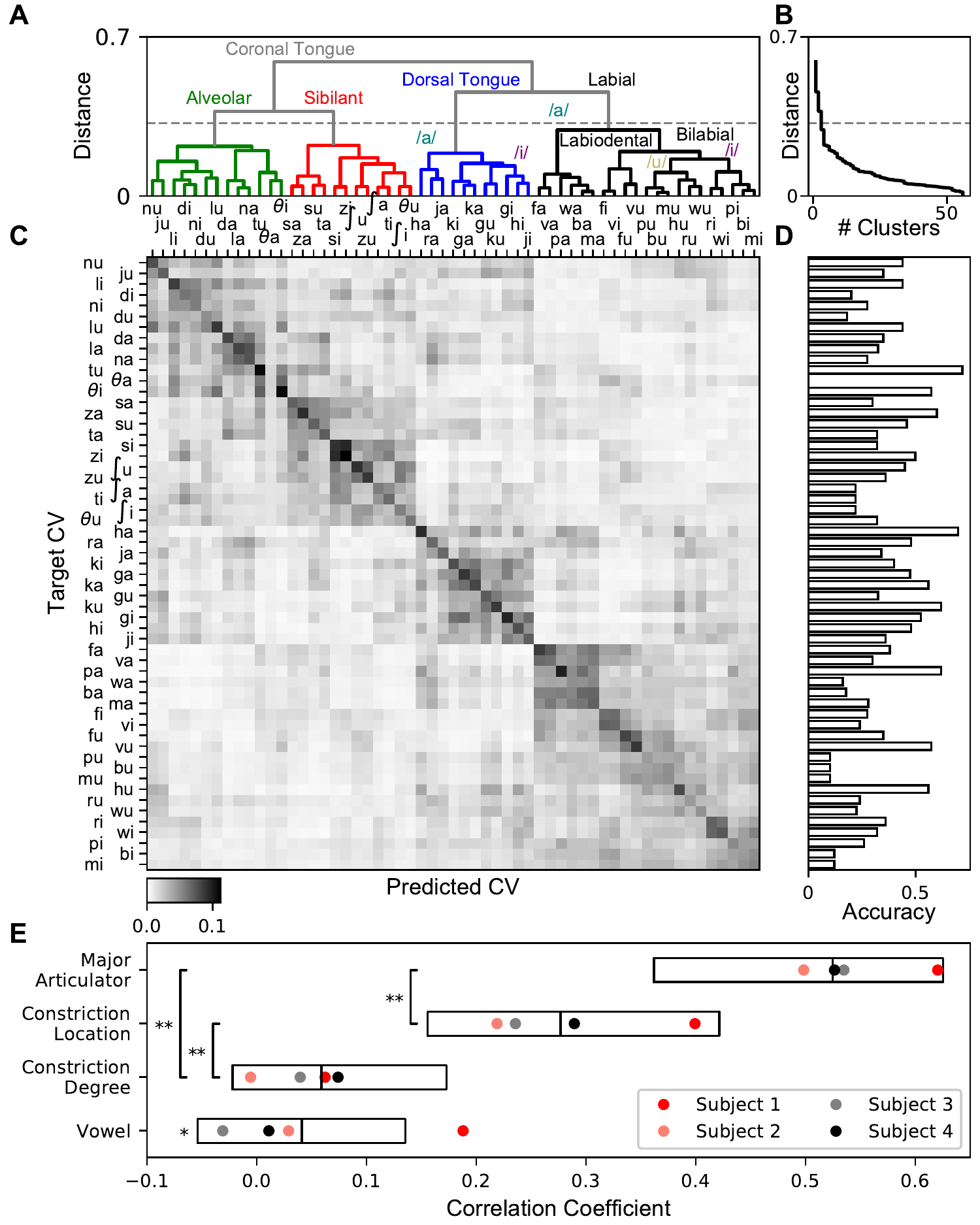}
	\end{center}
	\caption{{\bf Deep network predictions reveal a latent articulatory hierarchy from single-trial ECoG recordings.} \textbf{A} The dendrogram from a hierarchical clustering of deep network predictions on the test set from Subject 1. The threshold for the colored clusters (dashed gray) is determined from inspection of the number of clusters as a function of distance cutoff shown in \textbf{B}. Clusters centroids are labeled with articulatory features shared by leaf CVs. \textbf{B} Number of clusters (vertical axis) as a function of the minimum cutoff distance between cluster centroids (horizontal axis). \textbf{C} Average predicted probability per CV for Subject 1. CVs are ordered from clustering analysis in \textbf{A}. \textbf{D} Accuracy of individual CVs for Subject 1. \textbf{E} Correlation between pairwise distances in deep network similarity space from \textbf{C} compared to distances in an articulatory/phonetic feature space for Major Articulator, Consonant Constriction Location, Consonant Constriction Degree, and Vowel, aggregated across all subjects. Center bar is the median and boundaries are 50\% confidence intervals. Colored circles indicate subject medians. $^{**}p<1\times 10^{-10}$, WSRT, $^*p<1\times 10^{-4}$ t-test, both Bonferroni corrected with $n=4$.}
	\label{fig:hierarchy}
\end{figure}

The dendrogram resulting from agglomerative hierarchical clustering on the trial averaged output of the softmax of the deep network (i.e., before thresholding for classification) for Subject 1 shows clusters spread across scales (Fig~\ref{fig:hierarchy}A). A threshold was chosen by inspection of the number of clusters as a function of cutoff distance (Fig~\ref{fig:hierarchy}B) and used to color the highest levels of the hierarchy. At the highest level, syllables are confused only within the major articulator involved (lips, back tongue, or front tongue) in the syllable. This is followed by a characterization of the place of articulation within each articulator (bilabial, labio-dental, etc.). At the lowest level there seems to be a clustering across the consonant constriction degree and vowel categories that capture the general shape of the vocal tract in producing the syllable. When ordered by this clustering, the soft confusion matrix (Fig~\ref{fig:hierarchy}C) resulting from the average output of the final layer softmax shows block-diagonal structure corresponding to the articulatory hierarchy. There is a large amount of variation in the per-CV accuracies (Fig~\ref{fig:hierarchy}D).

This hierarchy can be quantified by comparing the space of deep network prediction probabilities  and the space of articulatory features associated with each CV. This comparison was made by correlating pairwise CV distances in these two features spaces across all pairs of CVs. The resulting structure of correlations is consistent with an articulatory organization in vSMC (Fig~\ref{fig:hierarchy}C). The major articulators feature distances are most correlated with the distances between CVs in deep network space, then consonant constriction location, and finally consonant constriction degree and vowel.

Together, these results show that deep networks trained to classify speech from H$\gamma$ activity are learning an articulatory latent structure from the neural data. Furthermore, this structure is in agreement with previous analyses of mean spatial patterns of activity at separate consonant and vowel time points~\cite{bouchard2013}. Together, these results demonstrate the capacity of deep networks to reveal underlying structure in single-trial neural recordings.

\subsection*{The high gamma and beta bands show a diversity of correlations across electrodes and CVs}

Complex behaviors, such as speech, involve the coordination of multiple articulators on fast timescales. These articulators are controlled by spatially distributed functional areas of cortex. Lower frequency oscillations have been proposed as a coordinating signal in cortex. Previous studies have reported movement- or event-related beta ($\beta$)-H$\gamma$ desynchronization or decorrelation \cite{crone1998a, crone1998b, engel2010}. The differential structure of these correlations across tasks and functions areas is not commonly analyzed. Since cortex often shows sparse and spatially-differentiated activity across tasks~\cite{bouchard2014a}, averaging over electrodes and tasks may obscure structure in the cross-frequency relationships.

\begin{figure}[!htbp]
	\begin{center}
		\includegraphics[width=5in]{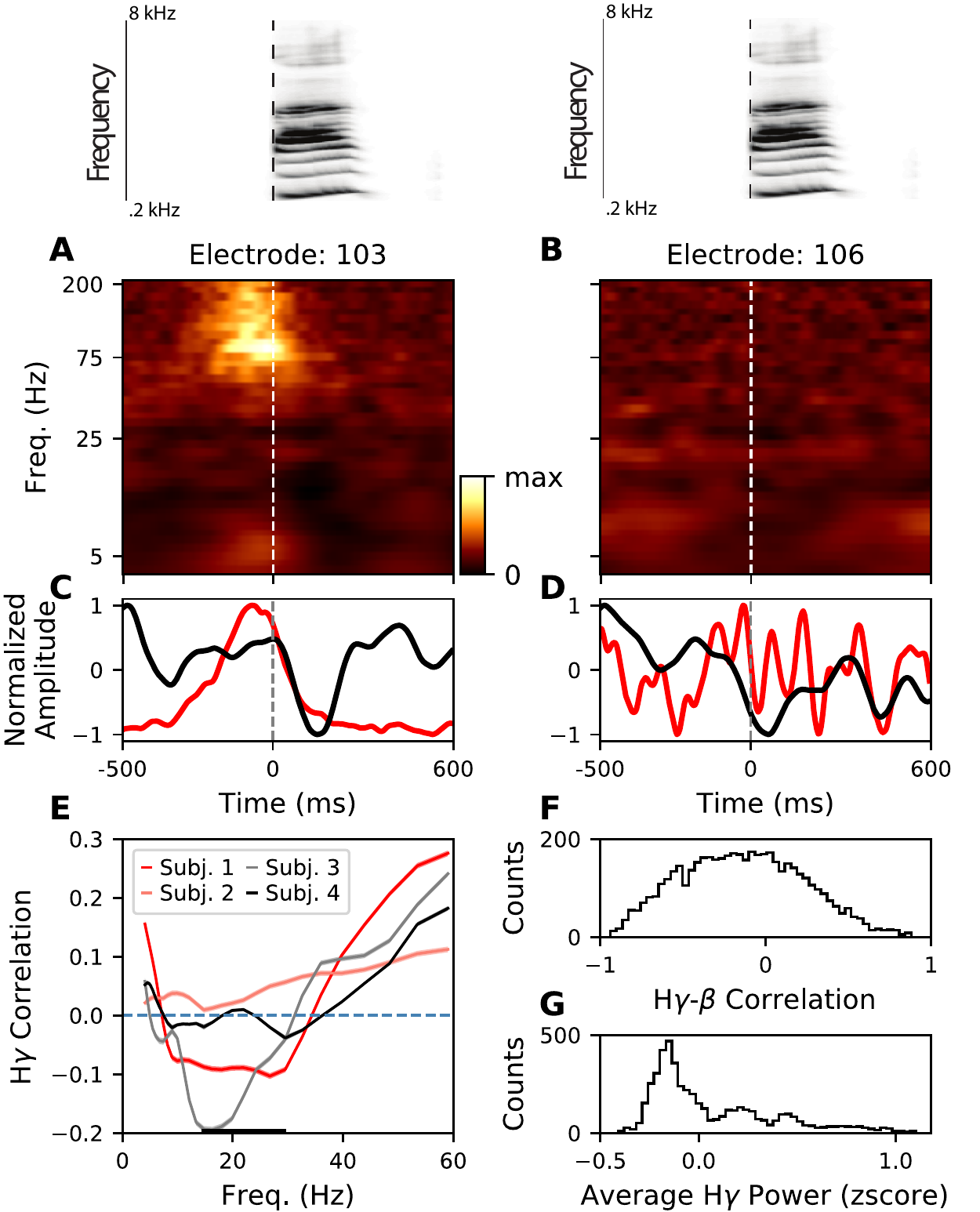}
	\end{center}
	\caption{{\bf H$\gamma$ and $\beta$ bands show diverse correlation structures across electrodes and CVs.} \textbf{A}-\textbf{B} Average amplitude as a function of frequency and time for an electrode with large activity during /ga/ production and for an electrode with no activity during /ga/ production. \textbf{C} and \textbf{D} Normalized (-1 to 1) H$\gamma$ (red) and $\beta$ (black) activity from \textbf{A} and \textbf{B} respectively. Non-trivial temporal relationships can be seen in \textbf{C} which are not apparent in \textbf{D}. \textbf{E} The average correlation ($\pm$ s.e.m.) between the H$\gamma$ amplitude and the single frequency amplitude is plotted as a function of frequency for each subject. Thickened region of the horizontal axis indicates the $\beta$ frequency range. \textbf{F} Histogram of the H$\gamma$-$\beta$ correlation coefficients for all CVs and electrodes for Subject 1. \textbf{G} Histogram of the z-scored H$\gamma$ power near the CV acoustic transition (time $=0$) for all CVs and electrodes for Subject 1.}
	\label{fig:xfreq1}
\end{figure}

The CV task and grid coverage allow average neural spectrograms (zscored amplitude as a function of frequency and time) to be measured at two electrodes during the production of the syllable {\textbackslash}ga{\textbackslash} (Fig~\ref{fig:xfreq1}A and B, median acoustic spectrogram is shown above). In order to investigate this, we measured cross frequency amplitude-amplitude coupling (correlation) for individual lower frequency bands and H$\gamma$. We also examine the aggregate $\beta$ band. Initially, we pool results across all electrodes and CVs in order to replicate methods from previous studies. The H$\gamma$ and $\beta$ amplitudes show a diverse set of temporal relationships in these regions (Fig~\ref{fig:xfreq1}C and D). Across frequencies, H$\gamma$ correlation is positive for low frequencies ($<15$Hz), then we see negative and near-zero correlations between H$\gamma$ and the $\beta$ range across subjects, and finally the correlation rises for the $\gamma$ range ($30-59$ Hz) as the frequencies approach H$\gamma$ (Fig~\ref{fig:xfreq1}E). However, these mean correlations mask a broad range of H$\gamma$-$\beta$ correlations (Fig~\ref{fig:xfreq1}F) across H$\gamma$ activity (across CVs and electrodes). This includes a large number of positive correlations. Similarly, although most of the amplitudes measured are smaller than baseline (Fig~\ref{fig:xfreq1}G), there is a long tail to amplitudes larger than baseline (above 0).

This diversity of correlations and amplitudes across CVs and electrodes indicates there is potentially substructure in the data that is being averaged over. This motivates a disaggregated analysis of the amplitude-amplitude correlations. Naively, one might expect to see different cross-frequency relationships in areas that are actively engaged in a task compared to area which are not engaged. The broad coverage of the ECoG grid and the diversity of articulatory movements across the consonants and vowels in the task allow us to investigate whether there is substructure in the amplitude-amplitude cross frequency correlations.

\begin{figure}[!htbp]
	\begin{center}
		\includegraphics[width=5in]{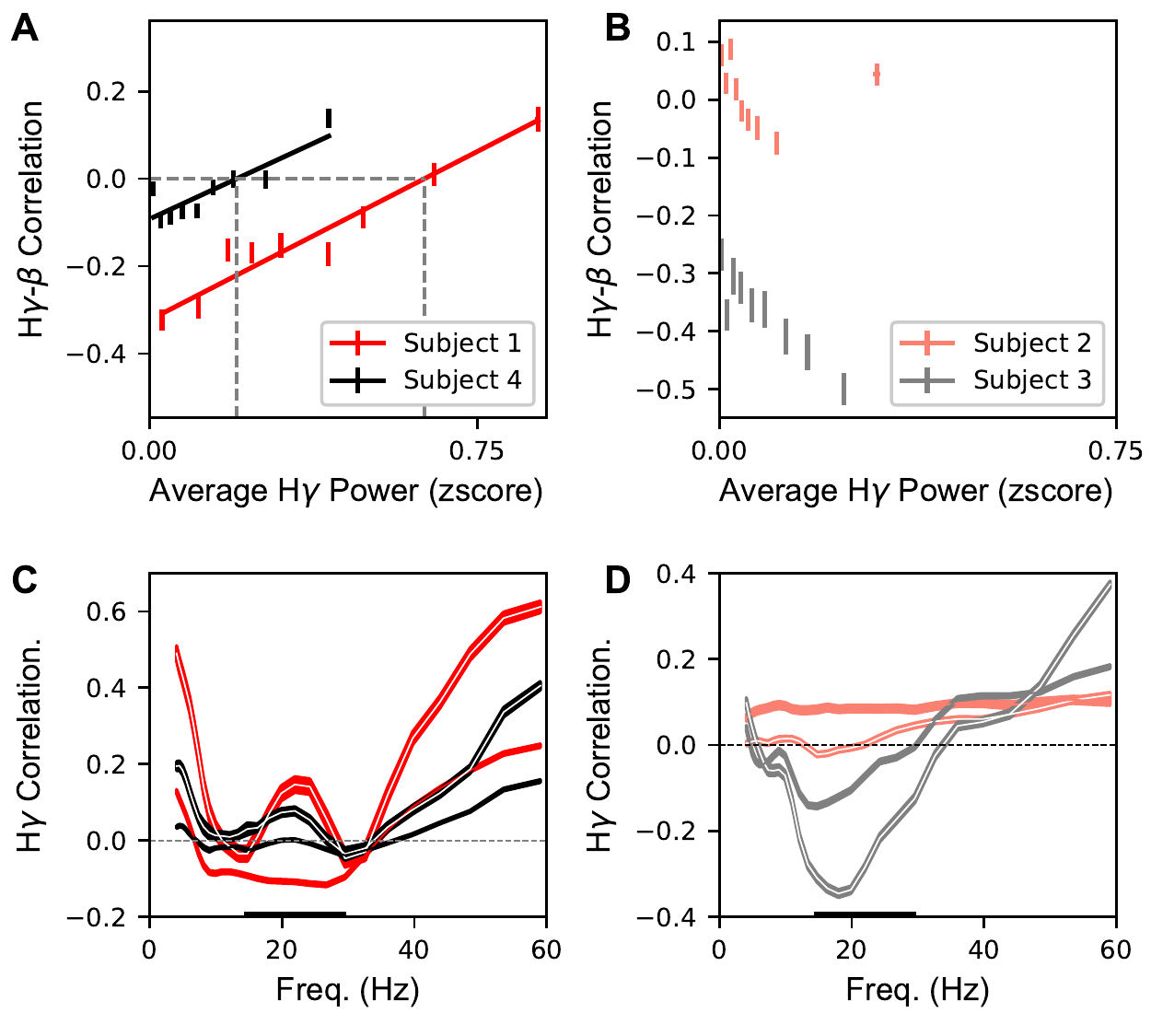}
	\end{center}
	\caption{{\bf H$\gamma$ and $\beta$ bands show positive correlations at active electrodes which are not found in inactive electrodes for subjects with high classification accuracy.} \textbf{A} The trial-averaged H$\gamma$-$\beta$ correlation coefficient across electrodes and CVs is plotted against the average H$\gamma$ power near the CV acoustic transition for Subjects 1 and 4. Solid lines indicate the linear regression fit to the data with positive z-scored amplitude. The vertical dashed gray line indicates the division in average H$\gamma$ power between `active' and `inactive' electrodes for subjects 1 and 4. Data is summarized in nine bins plotted ($\pm$ s.e.m.) per subject. \textbf{B} Same as \textbf{A}, but for Subjects 2 and 3, which have a much lower classification accuracy. \textbf{C} For the two subjects in \textbf{A}, the average ($\pm$ s.e.m.) correlation is plotted between the H$\gamma$ amplitude and the single frequency amplitude as a function of frequency separately for active (white center line) and inactive (solid color) electrodes. Thickened region of the horizontal axis indicates the $\beta$ frequency range. \textbf{D} Same as \textbf{C} for subjects in \textbf{B}.}
	\label{fig:xfreq2}
\end{figure}

In order to investigate this, we grouped the H$\gamma$ activity for each electrode and CV into ``active" and ``inactive" groups based on the average H$\gamma$ power and computed correlations for these two groups. For the two subjects with high accuracy, we observe a positive correlation between H$\gamma$ power and H$\gamma$-$\beta$ correlation (Fig~\ref{fig:xfreq2}A). For the two subjects with low CV classification accuracy, we observe a generally negative correlation between H$\gamma$ power and H$\gamma$-$\beta$ amplitude (Fig~\ref{fig:xfreq2}B).

The H$\gamma$ correlation can be recomputed separately for active and inactive electrodes per CV. For the subjects with high CV classification accuracy (Subjects 1 and 4), we find a novel signature of motor coordination in the active electrodes: a positive correlation in the $\beta$ frequency band (Fig~\ref{fig:xfreq2}C, lines with white centers). This is in contrast to the inactive electrodes, which show small or negative correlation (Fig~\ref{fig:xfreq2}C, solid lines) which is similar to the aggregated results (Fig~\ref{fig:xfreq1}E). For the two subjects with low CV classification accuracy (Subjects 2 and 3), the disaggregated results (Fig~\ref{fig:xfreq2}D) show less dichotomous structure.

Overall, we find that there is structure across bands in addition to cross-frequency relationship with the H$\gamma$ band which has been used in the preceding classification analysis. As far as we are aware, this is the first observation of dichotomous amplitude-amplitude cross frequency correlation during behavior. This observation was only possible because of the broad functional coverage of the ECoG grid and the diverse behaviors represented in the CV task.

\subsection*{Classification relevant information in lower frequency bands}

The gamma and H$\gamma$ band-passed CSEP amplitudes are commonly used both on their own and in conjunction with other frequency bands for decoding produced and perceived speech in humans due to their observed relation to motor and sensory tasks \cite{miller2007, mugler2014, bouchard2014a, yang2015, herff2015, leonard2016}. Other frequency bands have been shown to have amplitude or phase activity which is correlated with H$\gamma$ amplitude or spiking activity \cite{canolty2010, takahashi2015, rubino2006, richter2017}. Indeed, in the data used in this study, we find amplitude-amplitude correlation structure between H$\gamma$ and lower frequency bands. Although these correlations imply that information is shared between H$\gamma$ and other CSEP frequency bands, it is not known whether the other bands contain additional information about motor tasks beyond H$\gamma$ or whether the information is redundant.

In order to understand the relative information content in CSEP frequency bands, we classified CVs from two different sets of features. Linear classification methods would not give a satisfactory answer to this question since they are limited to simple hyper-plane segmentation of the data which may trivially lead to the result of no information. Indeed, since we have shown that deep networks can outperform linear methods when classifying from H$\gamma$, they are also candidates for showing whether there is any relevant information in these bands. For the theta, alpha, beta, high beta, and gamma bands, each band's features were first used for classification and then concatenated with the H$\gamma$ features and used for classification. The raw classification accuracy and improvement beyond H$\gamma$ are two measures that give insight into information content in the other bands.

\begin{figure}[!htbp]
	\begin{center}
		\includegraphics[width=5in]{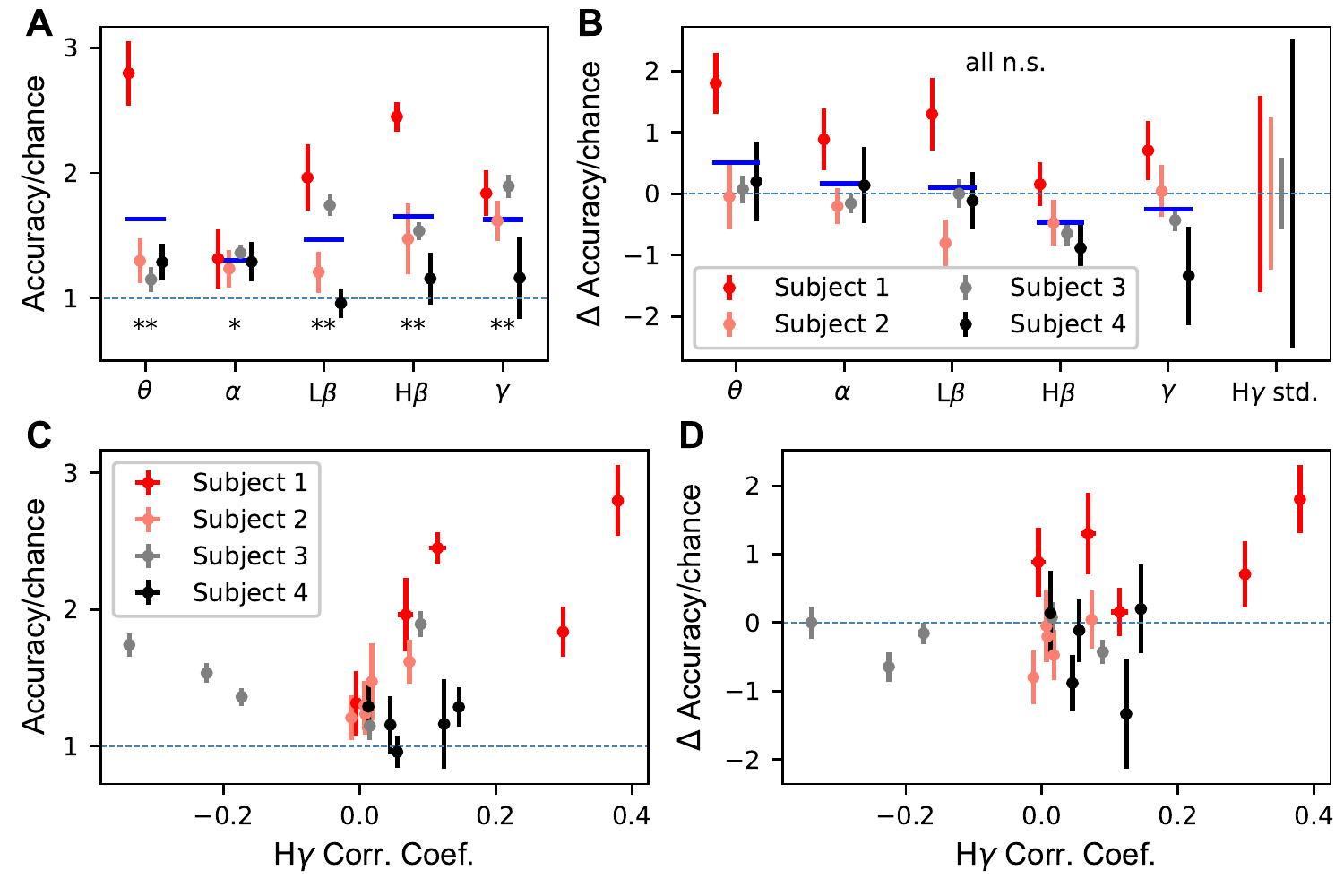}
	\end{center}
	\caption{{\bf Lower frequency bands do not contribute significant additional information to the CV classification task beyond H$\gamma$.} \textbf{A} The average accuracy ($\pm$ s.e.m., $n=10$) normalized to chance (chance = 1, dashed blue line) is shown for each frequency band and subject. Subjects are left-right jittered to avoid visual overlap. The solid blue line is the mean across subjects for a single band. \textbf{B} Average change in accuracy ($\pm$ s.e.m., $n=10$) from H$\gamma$ accuracy normalized to chance when band's features are concatenated with the H$\gamma$ features. The solid blue line is the mean across subjects for a single band. The H$\gamma$ accuracy cross-validation standard deviation ($n=10$) normalized to chance is plotted above and below zero in the right-most column for each subject for comparison. \textbf{C} Average accuracy ($\pm$ s.e.m., $n=10$) normalized to chance (dashed blue line, chance = 1) plotted against the correlation coefficient between H$\gamma$ and the lower frequency band for active electrodes for each band and subject. The blue dashed line indicates chance accuracy. \textbf{D} Change in accuracy from H$\gamma$ accuracy normalized to chance plotted against the correlation coefficient between H$\gamma$ and the lower frequency band for active electrodes  for each band and subject. The blue dashed line indicates no change in accuracy. $^{**}p<0.001$, $^{*}p<0.01$, WSRT, n.s., not significant. All Bonferroni corrected with $n=5$.}
	\label{fig:multiband}
\end{figure}

Fig~\ref{fig:multiband} shows the accuracies, normalized to multiples of chance, across the four subjects. Fig~\ref{fig:multiband}A shows the classification accuracies across subjects for single band features. Across subjects, all single bands have CV classification accuracies greater than chance, although subject-to-subject variation is observed. Although this is significantly above chance, the ranges of improvements for the subject means range between 1.5x to 2x chance, a small accuracies compared to H$\gamma$ accuracies which ranged from 6x to 21x chance. For the single band features, accuracy above chance implies that there is relevant information about the task in the bands. Fig~\ref{fig:multiband}B shows the chance in classification accuracy relative to H$\gamma$ accuracy, normalized to chance. No bands see a significant improvement in accuracy over the baseline accuracy obtained by classifying from H$\gamma$. Indeed, all measured mean changes in accuracy are smaller than the cross-validation standard deviations for the H$\gamma$ accuracy. Together, these results show that there is task-relevant information in lower frequency bands, but the information is largely redundant to the information contained in the H$\gamma$ amplitude.

The correlations observed in Fig~\ref{fig:xfreq1} and Fig~\ref{fig:xfreq2} imply that there is some shared information between the lower frequency bands and the H$\gamma$ band. However, the classification accuracies from H$\gamma$ alone (Fig~\ref{fig:accuracy}) are much higher than any other individual frequency band and are not improved by the addition of extra features from lower frequency bands. This shows that the high frequency CSEPs (H$\gamma$ band), which is commonly used in motor decoding, are highly informative signals.



\section*{Discussion}
The structure or information content of neural data is often estimated by regressing neural features against known features in the stimulus or behavior. Traditionally, this has been done with linear models, which are often poorly matched to the structure of this relationship. Here, we have shown that deep networks trained on high gamma (H$\gamma$) cortical surface electrical potentials (CSEPs) can classify produced speech with significantly higher accuracy than traditional linear or single layer models. Additionally, we found that deep networks have larger increases in performance as a function of training dataset size. As with many studies of human ECoG, there was substantial variability across subjects. While the precise nature of cross-subject variability is unknown, likely contributors are differences in SNR due to: i) uncontrollable variation in the degree of contact of electrodes with cortex, ii) differences in variance across recording sessions blocks, iii) degree of subject engagement in task. Further sources of variability could be lack of presence of particular functional representations in the recorded activity. Interestingly, we found no clear relationship with the number of samples, suggesting that the variance is due to issues with the underlying signal. When classifying syllables, deep networks achieved state-of-the-art accuracy and channel capacity: for the best subject, this was 38.3\% and 3.09 bits per syllable. At word durations from Mugler et al.~\cite{mugler2014} and one CV syllable per word duration, 3.09 bits per syllable corresponds to 5.9 bits per second or 59 word per minute~\cite{reed1998}. This could also be combined with a language model to improve accuracy in clinical applications~\cite{herff2015} towards the eventual goal of natural spoken speech rates (250-600 words per minute). Together, these results show that deep networks are a promising analytic platform for brain-computer interface (BCI) for speech prosthetics, an application where high accuracy and high training sample efficiency are crucial. Since deep networks are highly parameterized nonlinear models, their online interactions with learning may be more complex than typical methods~\cite{carmena2003}. Studying how they behave in an online BCI will be important step in integrating them into clinical settings.

We observed classification accuracies were highest, both relative to chance and linear models, for consonant-vowel syllables compared to the consonants or vowels individually. This is consistent with previous reports on the presence of both anticipatory and perseverative coarticulation effects in vSMC (see also Fig~\ref{fig:ecog_data}E)~\cite{bouchard2014, mugler2014}. Coarticulation refers to the fact that, at a behavioral level, the production of speech phonemes is systematically influenced by the surrounding phonemic context. For communication prosthetics, one might hope to decode the most atomic units, phonemes, and then express the combinatorial complexity of language through combinations of the small number of phonemes. Combined with other studies, the results presented here indicate that coarticulation is a feature of speech motor control that must be accounted for in BCIs.

In contrast to many commercial applications of deep learning, where optimizing prediction accuracy is often the primary goal, in science it is also desirable to extract latent structure from the data to advance understanding. In the context of the current study, we used deep networks to determine how features of speech production were extracted from the neural activity to solve the classification task. Examination of the consonant-vowel confusions made by the deep networks reveal the underlying articulatory organization of speech production in the vSMC.
At the highest level, the deep networks cluster the CVs into the major articulator involved in forming the consonant, i.e. lips, front tongue, or back tongue. The consonant constriction location, e.g. teeth-to-lips versus lips, is in the intermediate level of the hierarchy. Finally, consonant constriction degree and vowel are clustered at the lowest level of the hierarchy. Crucially, the consonant articulatory hierarchy is not present in the CV labels which means that the deep network is extracting this hierarchy from noisy, single-trial CSEPs during training. The articulatory organization we find is consistent with previous studies, which used PCA on the trial-averaged data at specific points in time~\cite{bouchard2013}. However, we note that, while consistent with previous findings, the hierarchy observed here reflects structure across consonants and vowels together. This could not have been examined with our previous methodology, which required analyses at separate time points. In this way, deep networks were able to extract novel, more general structure from the data, and did so with much less human supervision.

Previous studies of motor cortex have claimed the existence of ``beta-desynchronization" (most commonly a decrease in beta amplitude) during motor production~\cite{crone1998a, pfurtscheller1999}. This has led to a variety of hypothesized functions of beta ($\beta$) band in motor preparation and control, with little consensus across studies. A common methodology in many of these previous studies (especially those done in humans, where the number of samples is small and function of cortex is often sub-sampled) is to aggregate data across all electrodes and tasks. For the two subjects for which there was high-quality decoding accuracy, and thus, likely higher quality CSEP recordings, we found a novel positive coupling, i.e., correlation, between the $\beta$ band and the H$\gamma$ band amplitudes. The positive correlation was band-limited, occurring in the $\beta$ range with a peak near 23Hz, and present at electrode-syllable combinations in which the electrode was active. Thus, uncovering this correlation required that we disaggregate the relation between $\beta$ and H$\gamma$ according to whether an electrode, i.e., articulator, was engaged in the production of a given speech sound. The presence of this coupling is correlated with the classification accuracy from the H$\gamma$ amplitude across subjects. The coupling in engaged functional areas is an example of the possible pitfalls of aggregation across functional areas and specific behaviors or stimuli when the combination of spatial specialization of function and task structure gives rise to sparse activation patterns. 

Frequency bands besides H$\gamma$ are known to contain information about stimuli, behavioral, and state variables~\cite{crone1998a, crone1998b, pfurtscheller1999, rubino2006, miller2007, canolty2010, michalareas2016}. However, comparisons of task-relevant information are rarely made. Information theory provides a way of measuring the amount of information about a task in a neural signal, the mutual information, but measuring mutual information across continuous, high dimensional signals is notoriously difficult. In the context of classifying discrete speech tokens, this information can be approximated through the information transfer rate. Being able to compare information across features is particularly useful for CSEPs which results from a variety of electrical processes in the brain~\cite{buzsaki2012}. Since they achieved higher accuracy then linear or single layer methods, deep networks optimized for accuracy can put a tighter bound on the task-relevant information in a set of neural features. We found that, for frequency bands lower than H$\gamma$, we show that it is possible to decode speech syllables with above chance accuracy, thought at relatively modest levels. Furthermore, when combined with H$\gamma$ features, the relative improvement in accuracy above H$\gamma$ accuracy is small compared to the cross-validation noise. Thus, for BCI, these results imply that, for the CV task examined here, only H$\gamma$ activity (or higher frequency signals) need be acquired and analyzed: the other parts of the signal may profitably not be acquired to minimize data acquisition hardware and signal-processing in the decoder.

Although deep networks have shown the ability to maximize task performance across scientific and engineering fields, they are still largely black boxes \cite{shwartz2017}. While there has been some initial investigations~\cite{saxe2013, zeiler2014, li2015, nguyen2016, achille2017}, theoretical and empirical studies have not yet shown how deep networks disentangle the structure of a dataset during training. Currently, deep networks are most commonly used in science in cases where understanding of the deep network's hidden representation is not needed. While we have taken some initial steps in that direction by examining the networks confusions, revealing how the deep networks disentangled articulatory features will be an important extension of this work. In general, understanding the interaction between dataset structure and deep network training will make deep networks more broadly useful as a tool for data analytics in science.

Neuroscience continues to create devices to measure more features in the brain while the stimuli or behavior during data collection become more complex and naturalistic. As the complexity of datasets increase, the tools needed to disentangle and understand these datasets must also evolve. Recently, deep networks have shown promise in analyzing and modeling neural responses in this work and others~\cite{agarwal2015, mcintosh2016, benjamin2017}. Moving beyond their utility as high-accuracy regression methods will require a more profound understanding of how deep networks learn to represent complex structure from data sets, and tools to extract that structure so as to provide insights to humans. Indeed, many of the open theoretical and analytical challenges facing deep networks are also core to understanding the brain. 

%
%

\section*{Supporting information}

\paragraph*{S1 Appendix.}
\label{S1_Appendix}
{\bf Deep network hyperparameters.} Fully-connected (FC) deep networks were training using Pylearn2 and Theano~\cite{goodfellow2013, theano2016}. Hyperparameters are listed in Table \ref{table:hyp}. Nesterov momentum was used as an optimizer with fixed initial momentum fraction (0.5). The momentum fraction was linearly increased per epoch, starting after the first epoch, to its saturation value. The initial learning rate was exponentially decayed per epoch to a minimum value. Many float hyperparameters were searched in log-space since they typically range over a few orders of magnitude.
\begin{table}[h]
	\caption{Hyperparameters for deep networks}
	\label{table:hyp}
	\begin{center}
		\begin{tabular}{lll}
			\multicolumn{1}{c}{\textbf{Name}}  &\multicolumn{1}{c}{\textbf{Type}} &\multicolumn{1}{c}{\textbf{Range/Options}}
			\\ \hline \\
			Init. momentum & Float & .5 \\
			Terminate after no improvement epochs & Int & 10\\
			\\ \hline \\
			Num FC Layers & Int & 1 : 2 \\
			FC dim & Int & dim(task) : 1000 \\
			FC layer type & Enum & ReLU, Tanh, Sigmoid \\
			$\log_{10}$ Weight init. scale & Float & -5 : 0 \\
			$\log_{10}$ Learning rate init. & Float & -3 : -1 \\
			$\log_{10}$ Min. learning rate & Float & -5 : -1 \\
			$\log_{10}$ One-minus learning rate decay & Float & -5 : -1 \\
			$\log_{10}$ One-minus final momentum & Float & -2 : -3.0102e-1 \\
			Momentum saturation epoch & Int & 1 : 50 \\
			Batch size & Int & 15 : 256 \\
			Max epochs & Int & 10 : 100 \\
			One-minus input dropout rate & Float & 3.0e-1 : 1 \\
			Input dropout rescale & Float & 1 : 3 \\
			One-minus hidden dropout rate & Float & 3.0e-1 : 1 \\
			Hidden dropout rescale & Float & 1 : 3 \\
			$\log_{10}$ $L_2$ weight decay & Float & -7 : 0 \\
			Max filter norm & Float & 0 : 3 \\
		\end{tabular}
	\end{center}
\end{table}

\begin{figure}[!h]
		\begin{center}
			\includegraphics[width=7in]{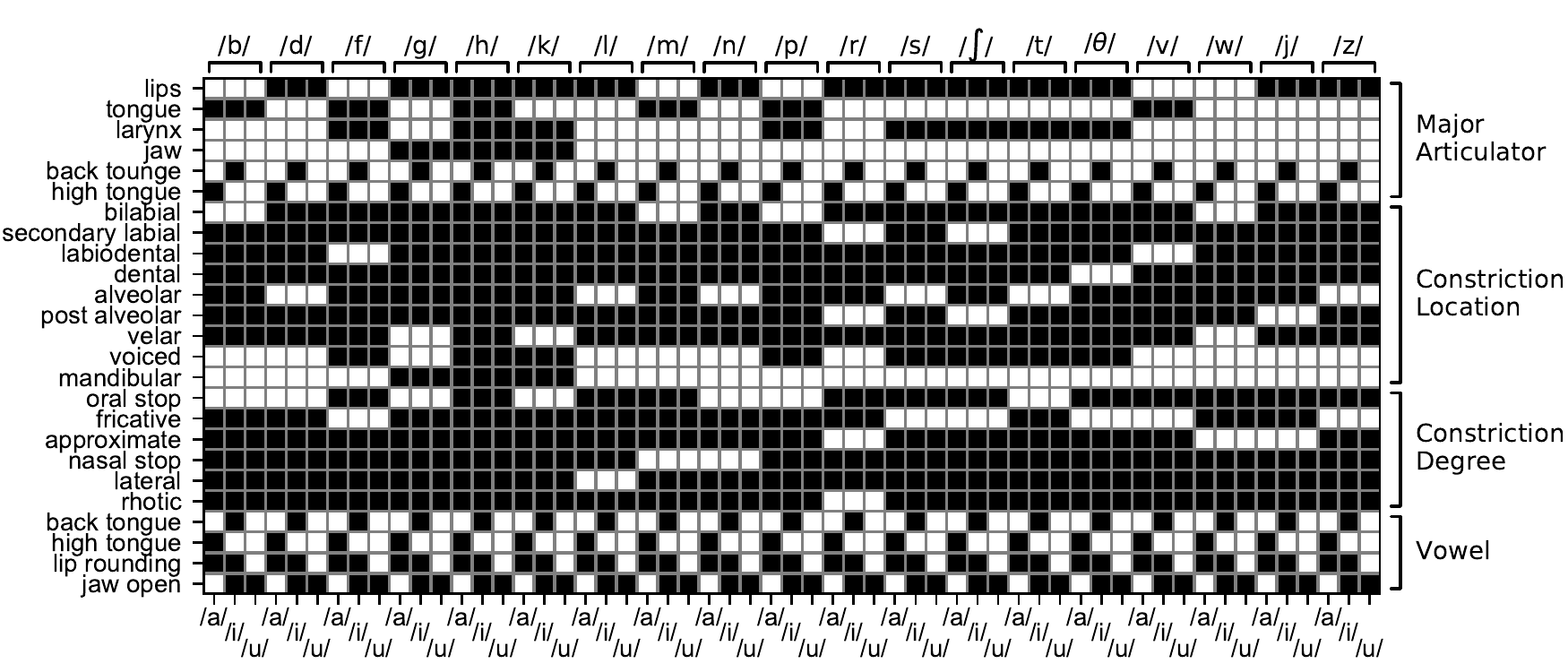}
		\end{center}
		\caption{{\bf (S1 Fig.) Articulatory features for comparison to deep network prediction features.} For each consonant vowel pair (labeled along top and bottom, respectively), a binary feature vector is shown (white indicates the presence of the feature). The grouping into major articulator, consonant constriction location, consonant constriction degree, and vowel features is shown on the right edge.}
		\label{fig:articulatory_features}
\end{figure}

\paragraph*{S1 Fig.}
\label{S1_Fig}
{\bf Articulatory features.} See Fig~\ref{fig:articulatory_features}.

\section*{Acknowledgments}
J.A.L. was funded by LBNL-internal LDRD “Neuromorphic Kalman Filters” led by Paolo Calafiura and LBNL-internal LDRD “Deep Learning for Science” led by Prabhat.
K.E.B. was funded by LBNL-internal LDRD “Neuro/Nano-Technology for BRAIN” led by Peter Denes, LBNL-internal LDRD “Neuromorphic Kalman Filters” led by Paolo Calafiura, and LBNL-internal LDRD “Deep Learning for Science” led by Prabhat. E.F.C. was funded by the US National Institutes of Health grants R00-NS065120, DP2-OD00862 and R01-DC012379, and the Ester A. and Joseph Klingenstein Foundation.
This project used resources of the National Energy Research Scientific Computing Center, which is supported by the Offce of Science of the U.S. Department of Energy under Contract No. DE-AC02-05CH11231.


%
%
%
\bibliographystyle{unsrt}
\bibliography{library}
%
%
%
%

\end{document}